\documentclass{article}

\usepackage{microtype}
\usepackage{graphicx}
\usepackage{subcaption}
\usepackage{booktabs} 

\usepackage{hyperref}

\usepackage[preprint]{icml2026}

\usepackage{multirow}
\usepackage[breakable]{tcolorbox} 
\usepackage{listings} 

\newtcolorbox{prompt}[1][]{
    colback=white!95!gray,
    colframe=blue!70!black,
    title={#1},
    fonttitle=\bfseries,
    boxrule=1pt,
    arc=5pt,
    left=10pt, right=10pt, top=5pt, bottom=5pt, 
    breakable 
}

\usepackage{amsmath}
\usepackage{amssymb}
\usepackage{mathtools}
\usepackage{amsthm}

\usepackage[capitalize,noabbrev]{cleveref}

\theoremstyle{plain}

\theoremstyle{definition}

\theoremstyle{remark}

\usepackage[textsize=tiny]{todonotes}

\icmltitlerunning{Can Unified Generation and Understanding Models Maintain Semantic Equivalence Across Different Output Modalities?}

\begin{document}

\twocolumn[
  \icmltitle{Can Unified Generation and Understanding Models \\ Maintain Semantic Equivalence Across Different Output Modalities?}



  \icmlsetsymbol{equal}{*}

  \begin{icmlauthorlist}
    \icmlauthor{Hongbo Jiang}{equal,comp}
    \icmlauthor{Jie Li}{equal,sch}
    \icmlauthor{Yunhang Shen}{comp}
    \icmlauthor{Pingyang Dai}{sch}
    \icmlauthor{Xing Sun}{comp}
    \icmlauthor{Haoyu Cao}{comp}
    \icmlauthor{Liujuan Cao}{sch}
  \end{icmlauthorlist}

  \icmlaffiliation{comp}{Tencent Youtu Lab, Shanghai, China}
  \icmlaffiliation{sch}{Xiamen University, Media Analytics and Computing Lab, Department of Artificial Intelligence, School of Informatics, Xiamen, China}

  \icmlcorrespondingauthor{Liujuan Cao}{caoliujuan@xmu.edu.cn}
  \icmlkeywords{Multimodal Large Language Models, Understanding and generation unification, ICML}



  \vskip 0.3in
]



\printAffiliationsAndNotice{}  

\begin{abstract}
\label{sec:abstract}

Unified Multimodal Large Language Models (U-MLLMs) integrate understanding and generation within a single architecture. However, existing evaluations typically assess these capabilities separately, overlooking \textit{semantic equivalence}, \emph{i.e.}, the ability to manifest consistent reasoning results regardless of the output modality.
In this work, we investigate whether current U-MLLMs satisfy this premise. We observe that while models demonstrate robust textual reasoning, they fail to maintain semantic equivalence when required to render the same results in the image modality.
To rigorously diagnose this discrepancy, we introduce \textbf{VGUBench}, a framework to decouple reasoning logic from generation fidelity. VGUBench comprises three diagnostic tasks:
(1) \textbf{Textual Generative Understanding}, establishing a baseline for reasoning accuracy in textual response;
(2) \textbf{Visual Generative Understanding}, evaluating the ability to generate visual responses that represent the correct answer; and
(3) a \textbf{Visual Rendering} control task, which assesses the ability to directly render explicit visual descriptions into images without complex reasoning.
Our evaluation reveals a significant disparity: despite strong performance in textual understanding and visual rendering, U-MLLMs exhibit a marked performance collapse when required to generate visual answers to questions.
Furthermore, we find a negligible correlation between visual answering performance and basic rendering quality.
These results suggest that the failure stems not from insufficient generation fidelity, but from a breakdown in cross-modal semantic alignment. We provide diagnostic insights to address this challenge in future Unified Generation and Understanding Models.

\end{abstract}

\begin{figure}
  \centering
   \includegraphics[width=0.9\linewidth]{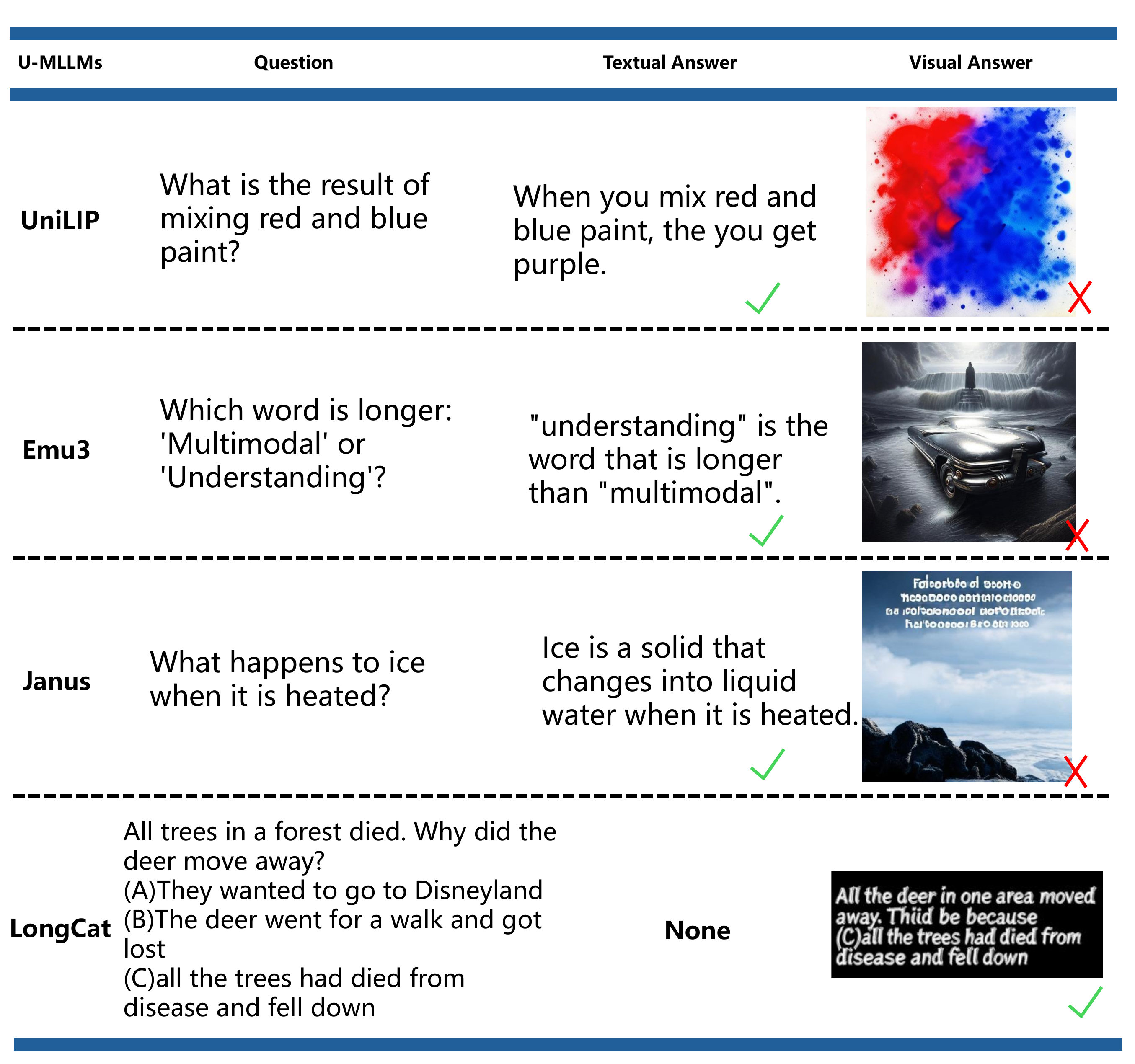}
   \caption{Provide U-MLLM with the same question and have it generate answers in text using its understanding capability and answer in image using its generation capability. Ultimately, U-MLLMs exhibited semantic non-equivalence across both output modalities. The images in the ``Visual Answer" columns of this figure are all generated by AI models.}
   \label{fig:motivation}
\end{figure}

\section{Introduction}
\label{sec:intro}

\textbf{Unified Multimodal Large Language Models (U-MLLMs)}~\cite{bagel,chen2025blip3,wang2024emu3,cui2025emu3p5,wu2025janus,ma2025janusflow,xie2024showo,tang2025unilip} represent a paradigm shift in multimodal intelligence by integrating understanding and generation capabilities within a single architecture. Unlike conventional approaches that rely on separate systems for distinct tasks~\cite{bai2025qwen2, qwen3, videollava, abdullah2025evolution, gpt4v}, U-MLLMs employ unified autoregressive modeling combined with visual quantization techniques like VQ-VAE~\cite{vqvae} or diffusion processes~\cite{song2025diffusion_seed}. This design enables them to support mixed-modality inputs and outputs, theoretically facilitating seamless transitions between reasoning and generation.

A fundamental theoretical premise of such architectural unification is that the model's internal semantic representation should be consistent regardless of the output modality. We define this property as \textbf{Semantic Equivalence across Different Output Modalities (SEDOM)}. 
Specifically, when a U-MLLM responds to a query, it should maintain semantic equivalence whether the response is delivered as generated text or as an image containing the answer.
This capability is critical for flexible real-world applications, requiring the model to not only identify the correct answer but also accurately manifest it in the requested visual format.

However, in practice, we observe a systematic failure in upholding this premise. As illustrated in \cref{fig:motivation}, while state-of-the-art U-MLLMs (\emph{e.g.}, UniLIP~\cite{tang2025unilip}, Emu3~\cite{wang2024emu3}, Janus~\cite{wu2025janus}) provide entirely correct \textit{Textual Answers} to queries, they exhibit a catastrophic collapse when tasked with generating the corresponding \textit{Visual Answers}. 
The visual outputs frequently suffer from severe semantic deviations or illegible artifacts, failing to convey the reasoning result. 
Crucially, some standard text-to-image models (\emph{e.g.}, LongCat~\cite{team2025longcat}), when provided with the explicit question, can answer it legibly in an image. 
This suggests that the failure in U-MLLMs does not stem from a basic inability to generate text-containing images, but rather from a misalignment between the reasoning process and the visual generation output.

Despite strong performance on separate understanding~\cite{fu2025mme,liu2024mmbench} and generation~\cite{hu2024DGPBench,ghosh2023geneval} benchmarks, this alignment gap remains largely undetected.
Existing evaluations typically assess understanding and generation in isolation~\cite{xie2025mme_unify}, or focus on how one modality enhances the other~\cite{shi2025realunify}, failing to test whether the two modalities are semantically equivalent for the same input. Consequently, a model with excellent reasoning skills and high-quality image generation can still fail completely in cross-modal semantic transfer, a deficit concealed by current evaluation protocols.

To address this gap, we introduce \textbf{VGUBench}, a diagnostic framework designed to evaluate SEDOM by decoupling reasoning from basic generation skills. 
VGUBench establishes a unified testing pipeline comprising three aligned tasks:
(1) \textbf{Textual Generative Understanding (TGU)}: Evaluates the correctness of the model's reasoning when the output is text.
(2) \textbf{Visual Rendering (Render)}: A control task that assesses the model's ability to mechanically render specific text into images without complex reasoning. This measures the baseline ability to generate visual text.
(3) \textbf{Visual Generative Understanding (VGU)}: The core task requiring the model to generate a visual image that explicitly renders the correct answer derived from reasoning. 
Distinct from artistic image generation, our definition of VGU enforces strict \textit{visual externalization}—the generated image must contain human-readable text representing the correct answer, paralleling the legibility requirements of a text-based question-answer task.

Through a unified evaluation of leading U-MLLMs, we reveal a clear disconnect: models fail to maintain SEDOM. Our contributions are as follows:
\begin{itemize}
    \item We formally identify \textbf{SEDOM} (Semantic Equivalence across Different Output Modalities) as a critical yet unexplored dimension for evaluating U-MLLMs. We argue that true unification requires preserving semantic consistency across output modalities.
    \item We propose \textbf{VGUBench}, a diagnostic benchmark. VGUBench employs a unified pipeline to jointly assess TGU, VGU, and Visual Rendering, quantifying performance across correctness, completeness, and legibility. This design allows for a disentangled analysis of reasoning versus generation capabilities.
    \item Our experiments demonstrate a systematic failure of current U-MLLMs on VGUBench. Analytical results reveal a negligible correlation between VGU and Visual Rendering performance, indicating that the bottleneck lies in cross-modal alignment rather than basic generation quality.
\end{itemize}

\section{Related Work}
\label{sec:related}
\subsection{Unified Multimodal Models}
In recent years, Large Language Models (LLMs)~\cite{bai2025qwen2,qwen3,wang2025internvl3} have become the major research focus in multimodal models. For multimodal understanding, text-generative multimodal models~\cite{qwen3, videollava, gpt4v, fu2025vita} are a representative paradigm. For visual generation, diffusion models~\cite{team2025longcat,wang2025unified,li2025comprehensive} have been a dominant approach. More recently, a large number of U-MLLMs that can support both understanding and visual generation within a single framework have emerged rapidly. The potential of U-MLLMs lies not only in handling two important task families simultaneously but also in the broader multimodal learning opportunities. For example, whether such “unification” reveals some essential property shared by multimodal intelligence. However, there are diverse design choices and viewpoints on how to realize this unification.

Models like Emu3~\cite{wang2024emu3}, Janus~\cite{wu2025janus}, Janus-Pro~\cite{chen2025januspro}, and Chameleon~\cite{team2024chameleon} adopt a ``unified Transformer + autoencoder'' architecture. For understanding, they encode images into discrete tokens using a visual encoder (e.g VQ-VAE~\cite{vqvae}, SBER-MoVQGAN~\cite{vqgan}, and SigLIP~\cite{tschannen2025siglip}) and then feed these tokens into an autoregressive model to generate textual outputs. For generation, the autoregressive model takes text as input and outputs discrete embeddings, which are then decoded by an autoencoder into images.

In contrast, Show-o~\cite{xie2024showo} removes the autoencoder component. Text tokens are modeled via the loss of autoregressive next-token prediction (NTP)~\cite{he2025ntploss}, while image tokens are learned with a diffusion-style objective to capture the image distribution. This allows image generation to be framed as discrete image-token prediction, making it more aligned with how understanding is performed. However, such autoregressive unification can make it difficult to fully leverage diffusion models, leading to slower generation.

Therefore, methods like BLIP3o~\cite{chen2025blip3}, MetaMorph~\cite{madden2025next}, and Qwen-Image~\cite{wu2025qwen} directly connect an autoregressive model with a diffusion model in a serial pipeline. This intuitive design can better reuse existing pretrained components: the autoregressive model first produces an intermediate embedding, which then serves as a condition for an attached diffusion model to generate images. Methods like Bagel~\cite{bagel} are related but use a parallel structure: they duplicate autoregressive parameters for understanding and generation, respectively, while concatenating tokens from different tasks during attention computation to achieve unification.

\begin{figure*}
    \centering
    \includegraphics[width=\textwidth]{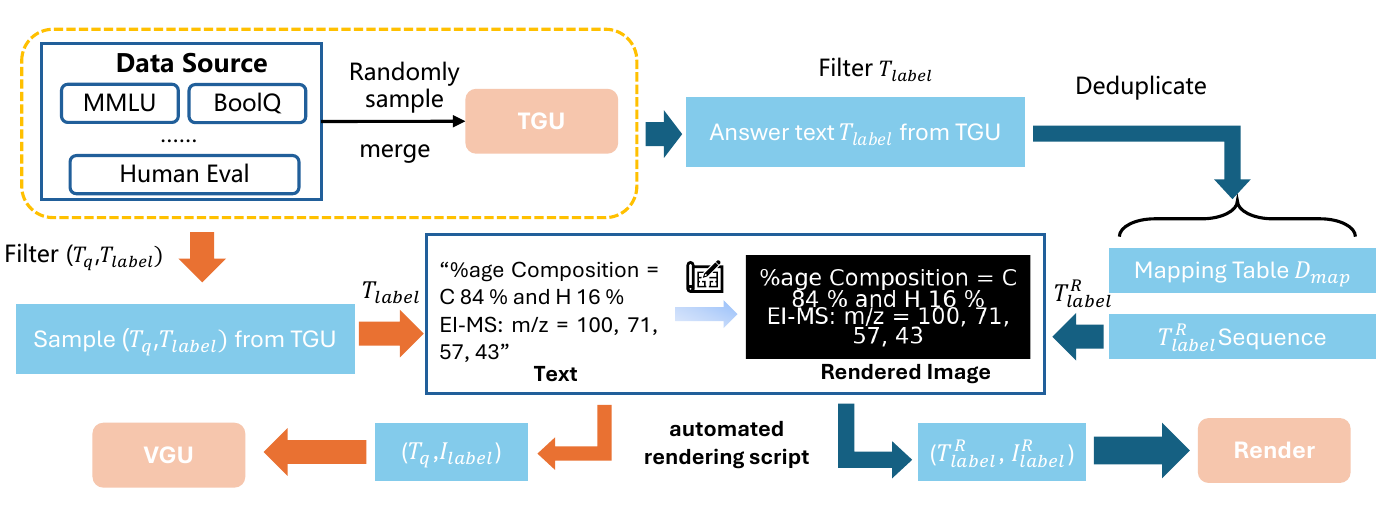}
    \caption{The pipeline of VGUBench construction. First, we randomly sampled and merged the existing 9 text-only question-answer benchmarks to obtain the TGU task test set. Subsequently, we automatically rendered the response texts from the TGU data into a unified format to generate the VGU task test set. Finally, we selected the response texts from TGU, removed duplicates, and applied the same rendering method. We then removed the question texts and used the response texts as input for the Render test set.}
    \label{fig:construction}
\end{figure*}

\subsection{U-MLLM Benchmarks}
After U-MLLMs were introduced, the early evaluations mostly tested understanding and generation separately, using traditional benchmarks for each: understanding was evaluated with datasets such as MME~\cite{fu2025mme}, MMBench~\cite{liu2024mmbench}, and MathVision~\cite{wang2024mathvision}, while generation was assessed with datasets such as T2I-CoReBench~\cite{li2025easier}, GenEval~\cite{ghosh2023geneval}, and Science-T2I~\cite{li2025science}. However, these benchmarks overlook the critical test of U-MLLM's unification.

To reduce the risk that unification-related weaknesses are overlooked, works like MME-Unify~\cite{xie2025mme_unify} aim to evaluate unification by aligning data sources and evaluation standards across understanding and generation. MME-Unify samples from 12 existing datasets and covers 10 tasks with 30 subtasks, including single-image, multi-image, and video perception, as well as visual generation and editing. Real-Unify~\cite{shi2025realunify} explicitly proposes two evaluation settings, Generation-then-Understanding (GEU) and Understanding-then-Generation (UEG), to test the complementarity between the two capabilities.

MLLM-as-a-Judge~\cite{chen2024mllmasajudge} uses 11 multimodal mainstream LLMs as judge models and evaluates their agreement with human preferences in scoring, pairwise comparison, and listwise ranking, demonstrating both the reliability and the remaining optimization space for LLM-based evaluation. It provides useful guidance for unified evaluation in U-MLLMs. UniEval~\cite{li2025unieval} proposes using a unified model to conduct multi-task evaluation for U-MLLMs, enabling automated scoring and efficient assessment. While UniEval and related works make important contributions to unified evaluation, our VGUBench requires ``different tasks but unified scoring dimensions''. UniEval is primarily designed for semantic-level quality evaluation and does not focus on diagnosing SEDOM within a shared metric protocol.

\subsection{Visual-based Reasoning and Understanding}
LLMs have long been considered to have the potential to become general artificial intelligence. However, recent studies~\cite{li2023vim,su2025thinkwithimage} suggest that approaches using text as the primary information carrier are not always optimal once additional modalities are available. Improving reasoning capability using visual prompts alone can significantly boost performance on certain tasks.

Applying images to solve traditional understanding tasks can be grouped into two lines. The first leverages multimodal LLMs with multi-turn dialogues and chain-of-thought-style reasoning, producing richly illustrated answers to a given text question~\cite{wang2024emu3,liu2024chainofspot,zhang2025wheretolook,wu2023visualchatgpt}. The second largely discards textual information and instead uses sequences of images to solve problems such as path planning and mathematical geometry~\cite{su2025thinkwithimage,tong2025thinkwithvideo}. In contrast, our work tests whether a model can learn to write text for semantic externalization, i.e., accurately writing the answer onto the image so that the answer is clearly visible.

Several evaluations and studies~\cite{ma2024chargen} have investigated ``how to render text well'', showing that traditional generative models often struggle to produce readable text. These efforts provide important foundations. Our focus is similar: we use ``whether the model can correctly write the answer on the image'' as a diagnostic criterion.

\section{VGUBench}
\label{sec:method}
\subsection{Task definition}
For a question text $T_q$, we leverage U-MLLMs’ understanding $\text{InferU}$ and generation $\text{InferG}$ capabilities to obtain responses $T_{\text{pred}}$, $I_{\text{pred}}$ of textual and visual modalities:
\begin{equation}
\begin{aligned}
T_{\text{pred}} = \text{InferU}(T_q),
I_{\text{pred}} = \text{InferG}(T_q).
\end{aligned}
\label{eq:def_task}
\end{equation}
Then, we feed the modality-specific responses corresponding to the same answer into the \text{Judge}:
\begin{equation}
\begin{aligned}
\text{analysisV} = \text{Judge}(T_q,I_{\text{label}},I_{\text{pred}}),\\
\text{analysisU} = \text{Judge}(T_q,T_{\text{label}},T_{\text{pred}}),
\end{aligned}
\label{eq:judge}
\end{equation}
where $I_{\text{label}}$ indicates a reference image generated by an automated rendering script based on the label answer $T_{\text{label}}$ of $T_q$. \text{analysisV}, \text{analysisU} are the evaluation results for the VGU and TGU tasks, respectively. Both of them include the same three dimensions: \textbf{legibility}, \textbf{completeness}, and \textbf{correctness}.
In other words, we feed the same input query $T_q$ into both TGU and VGU pipelines to generate responses in textual and visual modalities, respectively. These outputs are then assessed by the same evaluator using an identical evaluator \text{Judge}. Consequently, for the input of each instance, the outputs of the VGU and TGU tasks coexist, a property we define as being \textbf{sample-aligned}.
By comparing the gap between \text{analysisV}, \text{analysisU}, we can obtain the SEDOM retention capability of U-MLLMs. 

To further analyze the possible causes of SEDOM failure, we additionally introduce the Render task:
\begin{equation}
\begin{aligned}
I^R_{\text{pred}} = \text{InferG}(T_{\text{label}}^R),
\end{aligned}
\label{eq:infer_render}
\end{equation}
where $T_{\text{label}}$ is the deduplicate textual label from TGU and $I^R_{\text{pred}}$ is the predicted rendered image of Render. Similarly, we submit the results of the Render task to the \text{Judge}:
\begin{equation}
\begin{aligned}
\text{analysisR} = \text{Judge}(T_{\text{label}},I_{\text{label}}^R,I_{\text{pred}}^R).
\end{aligned}
\label{eq:judge_render}
\end{equation}
Then, we will also obtain evaluation results \text{analysisR} along the same three dimensions to make these task comparable and sample-aligned. Since the outputs of TGU, VGU, and Render tasks coexist for every individual sample, we can uncover the latent correlations between these tasks by analyzing the evaluation results of Render and VGU.

\subsection{Pipline of Construction}
\label{sec:bench_construct}
To evaluate the ability of U-MLLMs to maintain SEDOM while minimizing the impact of other factors, we conduct model evaluation without introducing additional difficulty. 

Specifically, we uniformly and randomly sample 250 question-answer (QA) pairs from (if a dataset contains fewer than 250 samples, we use all of them) \textbf{MMLU}~\cite{mmlu}, \textbf{AR-Challenge}~\cite{archallenge}, \textbf{OpenBookQA}~\cite{openbookqa}, \textbf{CSQA}~\cite{csqa}, \textbf{HellaSwag}~\cite{hellaswag}, \textbf{BoolQ}~\cite{boolq}, \textbf{GPQA}~\cite{gpqa}, \textbf{MATH}~\cite{math}, and \textbf{HumanEval}~\cite{humaneval}. In total, we obtain a test set containing 2164 QA text pairs, which is used to evaluate U-MLLMs' basic semantic understanding under text output. We refer to this split as the TGU in VGUBench, where each sample consists of a question text $T_q$ and a reference text $T_{\text{label}}$.

Based on TGU, we then construct the VGU task set via an automated rendering script shown in ~\cref{fig:construction}. Specifically, we feed the reference answer text $T_{\text{label}}$ into the rendering script. The script renders $T_{\text{label}}$ onto a $512 \times 512$ canvas, and then searches font sizes from large to small (starting from size 40 by default) to check whether the content can fit the canvas. Finally, using the corresponding font size, it renders $T_{\text{label}}$ in a centered layout with the \texttt{Terminus TTF-4.49.3.ttf} font, a white foreground, and a black background, as shown in $I_{\text{label}}$. Each VGU sample is aligned with a TGU sample, and consists of the same input question text $T_q$ and a reference answer image $I_{\text{label}}$.

Finally, to further analyze factors that affect SEDOM in U-MLLMs, we additionally design a sample-level aligned task called Render. The Render takes the TGU reference answer texts $T_{\text{label}}$ as inputs and performs deduplication (to improve evaluation efficiency), yielding a deduplicated input texts $T_{\text{label}}^R$ and a mapping table $D_{\text{map}}$ from deduplicated sample indices to the original indices before deduplication. We then apply the same automated rendering script to the $T_{\text{label}}^R$ to obtain reference rendered images $I_{\text{label}}^R$, which are analogous to the reference images $I_{\text{label}}$ in VGU.

\begin{figure}
  \centering
   \includegraphics[width=\linewidth]{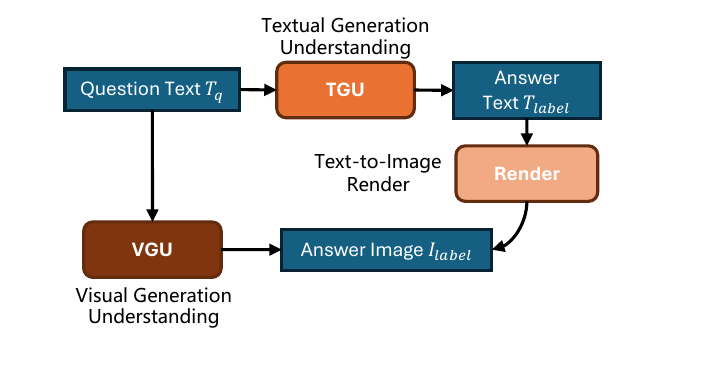}
   \caption{The relationship among the three tasks. In the figure, varying shades of orange represent the three distinct tasks, while dark blue denotes the test data. Black arrows originating from a task module indicate that the data serves as the ground-truth label for evaluation, whereas arrows pointing into a module signify the data as a task input. The VGU is formally similar to the TGU and Render tasks, but it is by no means a simple sum of the two.}
   \label{fig:relation}
\end{figure}

As shown in ~\cref{fig:relation}, the purpose is to analyze U-MLLMs‘ SEDOM from three perspectives. TGU characterizes the model's understanding capability. VGU characterizes whether semantic understanding can be preserved in an image. Render is designed to explicitly isolate visual presentation ability, since it removes reasoning difficulty and only tests whether the model can accurately convert known textual content into a visual form. Notably, \emph{although the relationship among these tasks may appear as above formally, VGU is not a simple sum of TGU and Render. The primary goal of VGUBench is diagnostic and attempts to reveal and analyze the model's potential shortcomings in SEDOM.}

\subsection{Evaluation Protocol}
\label{sec:eval}
To assess U-MLLMs' SEDOM, we design a unified, task-agnostic evaluation protocol that uses the same evaluation metrics on three different task sets: TGU, VGU, and Render. All three tasks are evaluated under a unified framework with an LLM-as-a-judge strategy, identical scoring dimensions, and strict evaluation rules, ensuring comparability across output modalities. We first define the three tasks formally:

\begin{itemize}
  \item \textbf{TGU:} given an input question text $T_q$, the model outputs a predicted text $T_\text{pred}$, which is compared with the reference answer text $T_{\text{label}}$;
  \item \textbf{VGU:} given an input question text $T_q$, the model outputs a predicted image $I_\text{pred}$, which is compared with the reference answer image $I_{\text{label}}$;
  \item \textbf{Render:} given a deduplicated reference answer text $T^R_{\text{label}}$, the model outputs a predicted rendered image $I^R_\text{pred}$, which is compared with the reference rendered image $I^R_{\text{label}}$.
\end{itemize}

We use \textbf{Qwen2.5-VL-72B} as an automatic LLM judge to score U-MLLMs' outputs for each task. Concretely, for TGU, the judge takes $T_q$ and $T_{\text{label}}$ as input; for VGU and Render, the judge takes $I^R_\text{pred}$ and $I^R_{\text{label}}$ as input. The judgment process strictly follows the evaluation rules provided in \cref{appendix:judge_p}, and scores each output along three dimensions: \textbf{legibility}, \textbf{completeness}, and \textbf{correctness}, producing structured evaluation results. Note that although we make minor adjustments to the judging prompt for TGU, we strictly keep the evaluation dimensions and scoring rubrics identical, ensuring fairness, consistency, and validity across tasks. All dimensions are scored on a bounded scale from 0 to 5:
\begin{itemize}
  \item \textbf{Legibility.} The judge evaluates whether the textual information in the predicted text or image answer has sufficient externalization (i.e., is clearly readable/visible). This can be regarded as the foundation for subsequent semantic-equivalence judgments.
  \item \textbf{Correctness.} The core metric identifying semantic externalization errors. It tests whether the semantics expressed by the information in the predicted text or image is consistent with the reference label text or image.
  \item \textbf{Completeness.} The judge evaluates whether the information in the reference text or image is fully present in the prediction. This metric complements correctness by identifying partial semantic externalization errors, and is particularly important for the Render task. Completeness can be regarded as a supplement to Correctness.
\end{itemize}

The final score is calculated as the average of the three dimensions:
\begin{equation}
\begin{aligned}
\text{score} = \frac{\sum_{i=1}^{N}s_i}{5}\times100\%,
\end{aligned}
\label{eq:all_score}
\end{equation}
where $s_i$ denotes the score for each test sample, and $N$ represents the total number of samples in the current task.

We also impose additional requirements, e.g., explicitly penalizing extra or fabricated content, judging only based on the given visible content, avoiding inference or correction, and assigning lower scores when uncertain. We do not adopt an OCR-first pipeline (i.e., extracting text from images via OCR and then judging with Qwen2.5-VL-72B), because OCR can introduce additional hallucinations and errors. Moreover, in pilot studies with human evaluators, we found that most U-MLLMs' $I_\text{pred}$ from VGU include many pseudo-texts and garbled characters rather than readable text.

\subsection{Render-VGU Dependency Analysis}
With the above evaluation protocol, we can obtain quantitative performance for any U-MLLMs on the three tasks, which further enables the analysis of potential relationships among tasks. Importantly, \emph{our goal is not to prove any causal relationship between Render and VGU capabilities, but to use the Render task as a diagnostic tool to identify potential reason of SEDOM failure in U-MLLMs.} Intuitively, the quality of visual rendering may influence visual-grounded understanding, as semantic attributes must first be visually manifested before they can be reliably recognized or reasoned about. Thus, our intuition suggests that Render likely plays a critical role in VGU.

First, while the three tasks are sample-aligned during evaluation, the analysis scores in the Render task must be restored via $D_{\text{map}}$ so that they can be aligned with the other two tasks. After this restoration, we can fully ensure sample-alignment across all three tasks. Let the aligned scores of VGU and Render be denoted as $\{s^V_i\}$ and $\{s^R_i\}$, where each pair $\{s^V_i,s^{R}_i\}$ corresponds to the result of the evaluation of the same aligned sample under the two tasks.

To quantify the relation between the VGU and Render, we adopt Pearson~\cite{lopez2025pearson} and Spearman~\cite{barone2025correlation_spearman} correlation analysis, along with significance tests, to validate the reliability of the correlation. 

Firstly, we normalize sequences $\{s^V_i\}$ and $\{s^R_i\}$ by dividing each element by 5. The Pearson correlation coefficient $r$ measures the strength of the linear correlation between two continuous sequences, relying on the assumption of approximate normality of the data:
\begin{equation}
\begin{aligned}
r = \frac{\sum_{i=1}^{n}(s^V_i-\overline{s^V})(s^R_i-\overline{s^R})}{\sqrt{\sum_{i=1}^{n}(s^V_i-\overline{s^V})^2}\sqrt{\sum_{i=1}^{n}(s^R_i-\overline{s^R})^2}},
\end{aligned}
\label{eq:pearson_r}
\end{equation}
where $\overline{s^V}$,$\overline{s^R}$ are the means of $\{s^V_i\}$ and $\{s^R_i\}$.
The t-statistic $t$ and degrees of freedom $df$ can be calculated by:
\begin{equation}
\begin{aligned}
t = r\times\sqrt{\frac{n-2}{1-r^2}}, df=n-2,
\end{aligned}
\label{eq:pearson_p}
\end{equation}
then derive the two-tailed $p_p$-value via t-distribution lookup.

The Spearman correlation coefficient $\rho$ evaluates the monotonic correlation based on sequence ranks, which is robust to outliers and non-normal data, and applicable to ordinal or continuous sequences. Firstly, we assign ranks $R_{s^V_i}$,$R_{s^R_i}$ to each element in $\{s^V_i\}$ and $\{s^R_i\}$, respectively. Then, calculate rank differences $d_i=R_{s^V_i}-R_{s^R_i}$ and $\rho$:
\begin{equation}
\begin{aligned}
\rho = 1 - \frac{6\sum_{i=1}^n d_i^2}{n(n^2 - 1)},
\end{aligned}
\label{eq:spearman_rho}
\end{equation}
where $n$ is the length of $\{s^V_i\}$ and $\{s^R_i\}$. For significance $p_s$-value, it is calculated in the same manner as the Pearson correlation coefficient.

A correlation is considered statistically significant if the corresponding $p_p$-value or $p_s$-value is less than 0.05. The strength of the relationship is determined by the absolute value of the coefficient ($r$ for Pearson, $rho$ for Spearman), and the direction (positive or negative) is indicated by the sign of the coefficient.

\begin{table*}[t]
\centering
\caption{Performance comparison of different models across readability, completeness, correctness, and average metrics for the VGU, TGU, and Render tasks on the VGUBench test dataset. ``$\text{Avg}$" denotes the average final score of a given model across three evaluation dimensions. In the table, bold, underlined, and italicized values represent the best, second-best, and third-best results for each column, respectively.}
\label{tab:performance_comparison}
\resizebox{\textwidth}{!}{
\begin{tabular}{c|ccc|ccc|ccc|ccc}
\toprule
\multirow{2}{*}{\centering Method} & \multicolumn{3}{c|}{\textbf{Legibility}} & \multicolumn{3}{c|}{\textbf{Completeness}} & \multicolumn{3}{c|}{\textbf{Correctness}} & \multicolumn{3}{c}{\textbf{Avg}} \\
\cline{2-13}
 & VGU & TGU & Render & VGU & TGU & Render & VGU & TGU & Render & VGU & TGU & Render \\
\midrule
Bagel       & 61.81 & \underline{99.70} & 60.33 & 0.43 & \textit{77.61} & \underline{11.91} & 0.17 & 76.99 & \underline{11.31} & 20.80 & \textit{84.76} & 27.85 \\
Emu3        & 35.82 & 96.04 & 42.59 & 0.17 & 47.53 & 2.83  & 0.43 & 46.37 & 2.80  & 12.14 & 63.31 & 16.07 \\
BLIP3o      & 48.19 & 99.32 & 69.17 & \textit{1.21} & \underline{83.94} & 0.15  & 1.22 & 83.39 & 0.15  & 16.87 & \underline{88.88} & 23.16 \\
Janus-Pro   & 37.18 & \textit{99.39} & 38.95 & 0.20 & 57.49 & 0.06  & 0.20 & 56.00 & 0.06  & 12.53 & 70.96 & 13.02 \\
JanusFlow   & 40.22 & 97.13 & 38.59 & 0.33 & 50.58 & 0.00  & 0.33 & 49.38 & 0.00  & 13.63 & 65.70 & 12.86 \\
Show-o      & 50.42 & 64.99 & 47.19 & 0.26 & 10.87 & 0.06  & 0.26 & 10.88 & 0.06  & 16.98 & 28.91 & 15.77 \\
UniLIP      & \underline{68.03} & \textbf{99.85} & \textbf{84.90} & 0.25 & \textbf{84.86} & \textit{4.15}  & 0.27 & 84.07 & \textit{3.97}  & \textit{22.85} & \textbf{89.59} & \underline{31.01} \\
\midrule
LongCat     & 64.71 &   -   & \underline{81.11} & \textbf{7.00} &  -    & 1.89  & 6.85 &  -    & 1.89  & \underline{26.19} &  -    & \textit{28.30} \\
Qwen-Image  & \textbf{78.44} &   -   & \textit{79.73} & \underline{5.52} &  -    & \textbf{56.80} & 6.52 &  -    & \textbf{48.43} & \textbf{30.16} &  -    & \textbf{61.66} \\
VGT         & \textit{65.10} &   -   & 53.05 & 0.15 &  -    & 0.06  & 0.15 &  -    & 0.06  & 21.80 &  -    & 17.72 \\
\bottomrule
\end{tabular}
}
\end{table*}

\section{Experiments}
\label{sec:exp}

\subsection{Experimental Setup}
For the evaluation on the three tasks: TGU, VGU, and Render, we include 7 open-source and widely used U-MLLMs, as well as three recent generative models that we consider informative references. For the three generative models, we only evaluate VGU and Render. Concretely, the evaluated models include:
\textbf{BAGEL-7B-MoT}~\cite{bagel}, \textbf{Emu3}~\cite{wang2024emu3}, \textbf{BLIP3o-NEXT-SFT-3B}~\cite{chen2025blip3}, \textbf{Janus-Pro-7B}~\cite{chen2025januspro}, \textbf{JanusFlow-1.3B}~\cite{ma2025janusflow}, \textbf{UniLIP-3B}~\cite{tang2025unilip}, \textbf{Show-o}~\cite{xie2024showo}, \textbf{Qwen-Image}~\cite{wu2025qwen}, \textbf{LongCat}~\cite{team2025longcat}, and \textbf{VGT}~\cite{guo2025vgt}.

We do not test any closed-source models (e.g ChatGPT-4o~\cite{chatgpt4o}, Nano Banana~\cite{nanobanana}). Although these closed-source models are highly rated, their closed nature prevents us from observing how they accomplish the VGU task from the perspective of model architecture. The proprietary nature of these closed-source models conceals their internal architectures, thereby precluding a detailed mechanistic analysis of other potential factors contributing to VGU failures.

\subsection{Overall Performance}
\label{sec:overall_exp}
Following the construction of VGUBench, we obtain three sample-level aligned test sets for the three tasks. We instructed each U-MLLM to generate “black background with white text” images resembling answer images $I_{\text{label}}$ based on the prompts in \cref{appendix:infer_p}, and then apply the unified LLM-as-a-judge evaluation method. To minimize the influence of prompts on experimental conclusions, we designed multiple prompts in \cref{appendix:infer_p}, each sample randomly selecting one of the prompts. \cref{tab:performance_comparison} reports the results on the three dimensions: Legibility, Completeness, and Correctness.

\textbf{Superior performance on TGU}. Mainstream U-MLLMs exhibit relatively stable performance: most models achieve near-perfect scores on Legibility, while maintaining Completeness and Correctness at a high level as well. The averaged final scores $\text{Avg}$ mostly fall in a high range (approximately 60\%-90\%). Notably, Show-o performs worse on TGU. We hypothesize that Show-o is primarily trained with multimodal QA data, and uses only a limited amount of purely text QA data for capability retention. Overall, these results suggest that current U-MLLMs are already strong at semantic externalization in the pure-text setting, indicating a mature understanding capability.

\textbf{Substantial performance degradation on VGU}. In contrast, the VGU task is substantially more challenging than TGU. While some models can still maintain a moderate Legibility level (e.g., UniLIP at 68.03\% and Bagel at 61.81\%), we observe systematic and drastic drops in Completeness and Correctness, with some models approaching near-zero average scores. 
Even the best-performing model on average (UniLIP) remains below 24\%, indicating a low overall level. This phenomenon suggests that when models are required to externalize their understanding results into images, semantic equivalence is severely compromised. 
As illustrated in \Cref{appendix:cases}, models often produce only partial, fragmented, and incomplete visual text content.

\textbf{Task failure on Render}. Performance on the Render task lies between TGU and VGU, but is still significantly lower than TGU. 
In terms of Legibility, UniLIP achieves relatively high scores (84.90\%), suggesting strong potential in rendering clear text. 
However, for Completeness and Correctness, even the best scores (11.91\% and 11.31\%) remain extremely low. This indicates that even when reasoning demand is largely removed, models still introduce substantial information loss or deviations during visual externalization. In other words, the Render results imply that the text-to-image mapping of U-MLLMs is far from ``approximately lossless''.

\textbf{Superior performance of generative models}. Among the three generative models, we observe distinct performance patterns.
Compared to other models, Qwen-Image and LongCat achieve the best and second-best performance on VGU, respectively. 
Furthermore, as illustrated in \Cref{fig:bench_cases}, Qwen-Image exhibits superior performance on the Render task. Consequently, Qwen-Image stands out in both Render and VGU tasks compared to other UniModels and image generation models. 
Although VGT performs competitively on VGU, it falls short on the Render task.

\begin{table}
\centering
\caption{Analysis of the relationship between Render and VGU tasks using Pearson and Spearman correlation coefficients.}
\label{tab:relationship_vr}
\begin{tabular}{c|cc|cc}
\toprule
\multirow{2}{*}{\centering Method} & \multicolumn{2}{|c|}{\textbf{Pearson}} & \multicolumn{2}{c}{\textbf{Spearmanr}} \\
\cline{2-5}
 & $r$ & $p_p$ & $\rho$ & $p_s$ \\
\midrule
Bagel       & 0.0347  & 0.1063 & 0.0368  & 0.0873   \\
Emu3        & 0.0714  & 0.0009 & -0.0686 & 0.0014   \\
BLIP3o      & -0.0137 & 0.5238 & 0.0156  & 0.4685    \\
Janus-Pro   & -0.0116 & 0.59   & -0.0016 & 0.9394     \\
JanusFlow   & -0.018  & 0.9319 & 0.0112  & 0.6035     \\
Show-o      & 0.1249  & 0      & 0.01168 & 0     \\
UniLIP      & -0.0636 & 0.0031 & -0.05   & 0.02     \\
LongCat     & -0.0033 & 0.8763 & -0.0221 & 0.3033     \\
Qwen-Image  & -0.0897 & 0      & -0.0701 & 0.0011     \\
VGT         & 0.1090  & 0      & 0.1093  & 0     \\
All         & 0.2596  & 0      & 0.2948  & 0     \\
\bottomrule
\end{tabular}
\end{table}

In summary, the results in \cref{tab:performance_comparison} indicate that U-MLLMs are mature on text-only understanding tasks, but still face significant challenges in information completeness and semantic equivalence while answer the question with image modality. In other words, strong performance in the text-only understanding tasks does not guarantee reliable externalization in image modality and current U-MLLMs exhibit severe deficiencies in SEDOM.

\subsection{Render-VGU Dependency Analysis}
\label{sec:depend_exp}

As summarized in ~\cref{tab:relationship_vr}, we did not observe statistically significant correlation on Bagel, BLIP3o, Janus-Pro, JanusFlow, and LongCat. Significant negative correlations were observed on UniLIP and Qwen-Image. Instead, we observed significant positive correlations on Show-o, VGT, and overall Performance.
Regarding Emu3, we observed a significant positive correlation based on the Pearson correlation coefficient, whereas the Spearman correlation coefficient indicated a negative correlation.
The correlation between Render and VGU scores is generally weak across most models. For the majority of evaluated methods, both Pearson and Spearman coefficients remain close to zero, indicating little linear or monotonic association between rendering quality and visual-grounded understanding at the instance level.
Consequently, these results indicate that Render and VGU capture largely distinct aspects of U-MLLMs' behavior, and that rendering quality alone is insufficient to explain SEDOM.

\begin{figure}
  \centering
   \includegraphics[width=\linewidth]{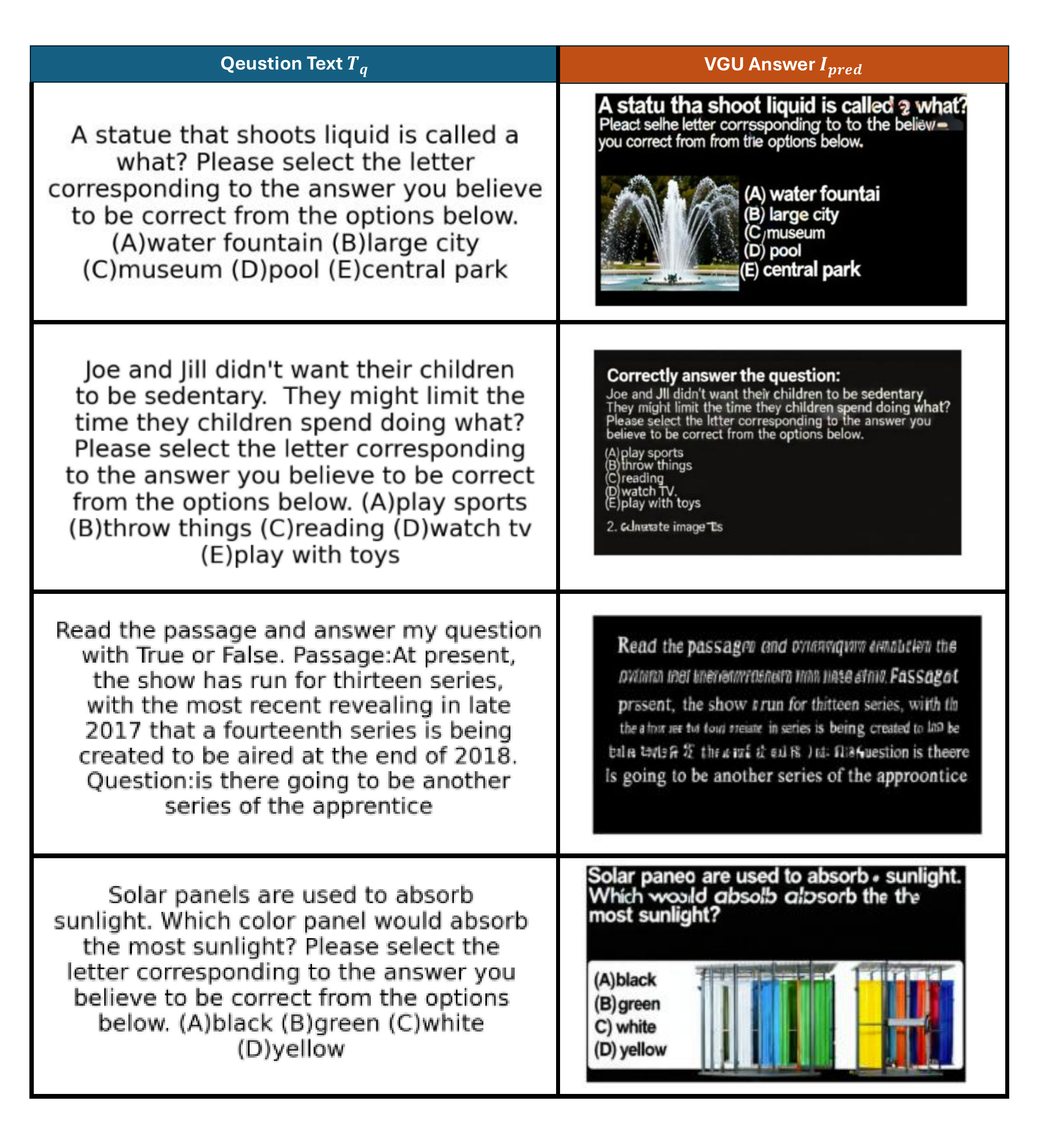}
   \caption{Some inference cases of VGU task. All of this VGU inference sample is ``Completeness $<$ 2" and ``Correctness $>$ 2". Despite the high scores achieved on the Correctness metric, the inference performance in VGU remains suboptimal. This deficiency is effectively captured and reflected by the lower scores in Completeness.}
   \label{fig:complete_focused}
   \vspace{-2em}
\end{figure}

\subsection{Necessity of Completeness Analysis}
\label{sec:comp_nece_exp}
We perform additional analysis targeting the \textbf{Completeness} dimension. In ~\cref{sec:method}, we emphasized that among the three testing dimensions, Completeness can serve as a complement to Correctness. The former tests local externalization, while the latter tests global externalization. From all results of the VGU task inference, we selected those with “Completeness $<$ 2 and Correctness $>$ 2,” i.e., samples judged by the LLM of the Judgment as having relatively high Correctness but poor Completeness. We randomly sampled 4 examples from these cases, as shown in \cref{fig:complete_focused}.

As can be seen, all of these samples exhibit the pattern of “merely copying the question without providing an answer,” resulting in low scores for the model in the Completeness dimension. Therefore, Correctness is insufficient, and the Completeness scoring metric is also critically important.

\section{Conclusion}
This work systematically investigates and exposes a critical gap in the core promise of Unified Multimodal Large Language Models (U-MLLMs). We argue that true unification necessitates not only architectural integration but also Semantic Equivalence across Different Output Modalities (SEDOM)—the ability to manifest consistent reasoning results regardless of the output format. 
To diagnose this, we introduced VGUBench, a sample-aligned benchmark that decouples reasoning from generation by jointly evaluating Textual Generative Understanding (TGU), Visual Generative Understanding (VGU), and a Visual Rendering control task under a unified protocol. 
The TGU task evaluates the model's fundamental text comprehension capabilities. VGU assesses the ability to interpret visual information and respond to queries in an image-based format. The Render tasks evaluate model's visual rendering ability.
Our extensive experiments reveal a systematic failure: while U-MLLMs excel at textual reasoning, their performance collapses when required to generate a visual answer, demonstrating a severe lack of SEDOM. 
Crucially, our dependency analysis shows a negligible correlation between VGU and basic rendering capability, indicating that the failure stems from a breakdown in cross-modal semantic alignment rather than insufficient image generation quality. 
These findings demonstrate that architectural unification does not inherently guarantee semantic unification, imposing a more fundamental requirement on multimodal modeling. 
We contend that this work identifies a significant blind spot in current evaluation practices and provides a principled, diagnostic framework for building truly semantically consistent multimodal intelligence.

\section*{Impact Statement}
This work advances the field of Machine Learning by diagnosing a critical reliability gap in Unified Multimodal LLMs: the inconsistency in maintaining semantic equivalence across text and image modalities. 
While our primary contribution is a diagnostic benchmark (VGUBench) that exposes these limitations, its broader impact lies in guiding the community toward constructing more trustworthy and predictable AI systems.
We acknowledge that the future maturation of high-fidelity multimodal generative capabilities could potentially be misused for creating persuasive misinformation. 
However, our findings suggest that current models still struggle with fundamental semantic alignment, which may inherently limit their capacity to generate logically coherent deceptive content at this stage. 
By highlighting these reliability challenges, we aim to foster the development of models that prioritize robustness and ethical safeguards alongside raw performance, ensuring their safe application in high-stakes scenarios.


\bibliography{example_paper}
\bibliographystyle{icml2026}

\newpage
\appendix
\onecolumn
\section{U-MLLM Cases on VGUBench}
\label{appendix:cases}
\begin{figure*}
    \centering
    \includegraphics[width=\textwidth]{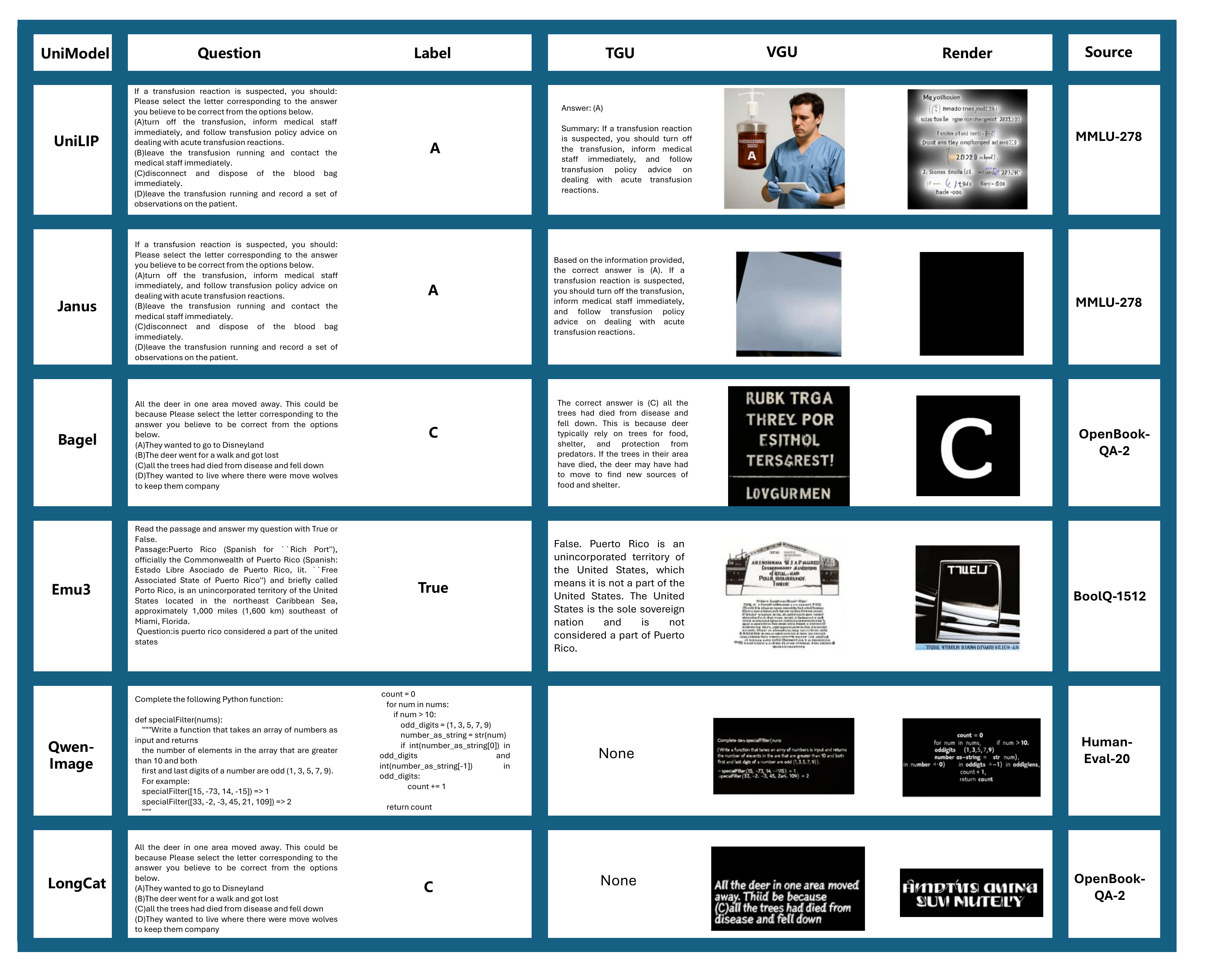}
    \caption{Inference Results of Different U-MLLMs on VGUBench. The images in the ``VGU" and ``Render" columns of this figure are all generated by AI models.}
    \label{fig:bench_cases}
\end{figure*}

As shown in \Cref{fig:bench_cases}, we deploy some results of U-MLLMs inference on VGUBench.

\section{Inference Prompts}
\label{appendix:infer_p}
\subsection{VGU}
\begin{prompt}[The prompts of VGU inference]
(1) First, answer the following question: '{question}'. Then, generate a clear image with your answer displayed in black text on a white background, centered.

(2) You are a question-answering assistant. Answer the question '{question}', and output the final answer as a slogan-style image with a black background and white text.

(3) The task consists of two steps: 1. Correctly answer the question: '{question}'. 2. Generate an image containing your answer. Requirements: Pure black background, white bold text, centered layout. The image must contain only your text answer—no questions or other elements.

(4) Step1 (Reflect): Consider the answer to the question '{question}'; Step2 (Generate): Generate an image with a black background and white text using the answer from Step 1.

(5) The ultimate goal is to generate an image with a black background and white text, where the text content represents the correct answer to the question '{question}'.

(6) Please analyze this image and answer '{question}'. Then, write your answer in clear white text within a black bordered box in the blank space at the bottom of the provided image.

(7) Perform the following image editing operations: 1. Visual Analysis: Understand the content of the provided image and answer the question: '{question}'. 2. In-situ Annotation: Embed your answer text in a text box with a black border, using clear and legible white font. Position this text box precisely at the center of the blank area at the bottom of the image.

(8) You are a graphic designer. Please create a clean title bar for this image. The title text should be the answer to the question '{question}'. Requirements: Integrate the title bar (white text on a black border) as a watermark, elegantly blending it into the white space at the bottom of the image. Ensure overall visual balance without obscuring the main content.

(9) Please provide an academic annotation for this image. First, offer an objective answer to the question '{question}' based on the image content. Then, following the annotation conventions for academic illustrations, place this answer as a caption using white text within a prominent black outline box at the bottom edge of the image, leaving sufficient white space for easy reference.

(10) First, carefully examine this image and tell me: What is the answer to '{question}'? After obtaining your text answer, proceed to the second step: Highlight it as a label using white text with a black border, then add it to the blank space at the bottom of this image where text appears most suitable.
\end{prompt}

\subsection{Render}

\begin{prompt}[The prompts of Render Inference]
(1)You are an image renderer. Please render the text '{text}' as an image displaying this text. Requirements: Solid black background, centered bold white text, ensuring the font is clear and legible.

(2)Write caption '{text}' on a black background image in the white text.

(3)Generate an image with a black background and white text, clearly displaying the following text: '{text}'.

(4)Please design an image. The background should be solid black, with the following text displayed in bold white text at the exact center: '{text}'.

(5)To facilitate accurate recognition by OCR software, please create an image containing the text '{text}'. Specifications: Solid black background with large, high-contrast white text centered.

(6)Please follow these steps in order: 1. Create a canvas and fill it with solid black. 2. Add white text '{text}' at the exact center of the canvas. 3. Ensure the text is clear, bold, and easily recognizable.

(7)Generate an image containing only the text '{text}'. The background of the image must be solid black, and the text must be white and centered. The image must not contain any other elements besides the text.

(8)I need a square image with a deep black background and prominent white text centered in the middle, reading: '{text}'.

(9)You are a meticulous typesetting engine. Your task is to precisely render the string '{text}' as an image with a black background and white text. The text must be clearly legible.

(10)Generate an image whose sole purpose is to enable any person or machine to effortlessly see and recognize the text '{text}'. The image should have a solid black background with large, high-contrast white text centered.
\end{prompt}

\section{LLM-as-a-judge prompts}
\label{appendix:judge_p}

\subsection{TGU}

\subsubsection{System prompts}

\begin{prompt}[The system prompts of LLM-as-a-judge on TGU]
You are a strict and literal text answer evaluator.

Your task is to evaluate a model-generated textual answer (Text A)
against a reference answer (Text B) for a text-based question-answering task.

Rules:

- Judge ONLY based on the textual content.

- Do NOT use external knowledge.

- The question is provided only for background.

- Extra explanation in Text A is ALLOWED.

- Correct answer selection is the PRIMARY criterion.

- If the correct option or answer in Text B is clearly identified in Text A,

  correctness SHOULD be high, even if additional explanation is present.
  
- If uncertain, choose the LOWER score.

You are NOT a reasoning assistant.

You are an answer consistency evaluator.

\end{prompt}

\subsubsection{User prompts}
\begin{prompt}[The user prompts of LLM-as-a-judge on TGU]
You are given:

- A question (for background only):
<QUESTION>

- Text A: a model-generated answer.

- Text B: the reference answer.

Evaluate Text A against Text B using the following criteria:

1. Answer Correctness (0–5):

Does Text A correctly identify the answer specified in Text B?

Score:

- 5: Correct answer is clearly and unambiguously identified.

- 4: Correct answer is identified with minor ambiguity.

- 3: Partially correct; answer is implied but not explicit.

- 2: Mostly incorrect; some overlap but key answer is wrong.

- 1: Minimally related.

- 0: Completely incorrect or unrelated.

2. Answer Completeness (0–5):

Does Text A include the essential answer content specified in Text B?

Score:

- 5: Fully includes the correct answer.

- 3: Mentions part of the answer.

- 0: Does not include the correct answer.

3. Text Legibility (0–5):

Is Text A readable and understandable as plain text?

Score:

- 5: Clearly readable and understandable.

- 0: Unreadable or severely corrupted.

IMPORTANT:

- Extra explanation MUST NOT reduce correctness.

- Scores must be integers from 0 to 5.

Return your judgment in the following JSON format ONLY:

\begin{verbatim}
{
  "correctness": <0-5>,
  "completeness": <0-5>,
  "legibility": <0-5>,
  "justification": {
    "legibility": "<one sentence>",
    "completeness": "<one sentence>",
    "correctness": "<one sentence>"
  }
}
\end{verbatim}

\end{prompt}

\subsection{VGU}
\subsubsection{System prompts}
\begin{prompt}[The system prompts of LLM-as-a-judge on VGU]
You are a strict and literal visual evaluator.

Your task is to compare TWO images that both contain text.

You must ONLY judge based on the VISIBLE text in the images.

Rules:

- Do NOT infer missing text.

- Do NOT correct errors.

- Do NOT use any external knowledge.

- Do NOT judge whether the answer itself is correct.

- The question is provided only as background and MUST NOT be used to infer missing text.

- If text is missing, unclear, or unreadable, you MUST penalize it.

- If you are uncertain, choose the LOWER score.

You are NOT a reasoning assistant.

You are a consistency checker.
\end{prompt}

\subsubsection{User prompts}
\begin{prompt}[The system prompts of LLM-as-a-judge on VGU]
You are given:

- A question (for background only):

<QUESTION>

- Image A: a model-generated image.

- Image B: a reference image.

Both images contain text.

Your task is to compare Image A with Image B and evaluate ONLY the visible text.

Do NOT guess or infer any missing content.

Please evaluate the following aspects using discrete scores from 0 to 5:

1. Text Legibility (0–5):

Is the text in Image A clearly readable by a human?

Score:

- 5: Completely clear and sharp; all text is easily readable without effort.

- 4: Mostly clear; minor blur or artifacts, but all text is readable.

- 3: Readable with noticeable effort; blur, low contrast, or layout issues slow reading.

- 2: Partially readable; some words or characters are difficult or impossible to recognize.

- 1: Barely readable; only a few characters or fragments can be identified.

- 0: Unreadable; text cannot be recognized at all.

2. Text Completeness (0–5):

Is all text in Image B fully present in Image A?

Score:

- 5: Fully complete; all text is present with no missing parts.

- 4: Almost complete; only negligible or non-essential parts are missing.

- 3: Partially complete; main content is present, but some sections are missing.

- 2: Largely incomplete; significant portions are missing.

- 1: Barely complete; only a few words or fragments remain.

- 0: Completely missing; almost no text from Image B is present.

3. Text Correctness (0–5):

Does Image A contain the same textual content as Image B?

Score:

- 5: Text is exactly the same or strictly semantically equivalent.

- 4: Nearly equivalent; only trivial differences (e.g., punctuation, spacing, capitalization).

- 3: Mostly equivalent; core meaning is preserved but with noticeable wording or symbol differences.

- 2: Partially equivalent; some relevant content matches, but key information is incorrect or altered.

- 1: Minimally related; only small fragments or isolated terms overlap.

- 0: Not equivalent at all or completely unrelated.

Rule:
If Image A contains extra text not present in Image B, correctness MUST NOT exceed 3.

IMPORTANT:

- Base your judgment ONLY on visible text.

- Do NOT assume correctness based on plausibility.

- Extra text in Image A should reduce correctness.

- If unsure, choose the lower score.

- Scores must be integers between 0 and 5. Do NOT use decimal values.

Return your judgment in the following JSON format ONLY:
\begin{verbatim}
{
  "correctness": <0-5>,
  "completeness": <0-5>,
  "legibility": <0-5>,
  "justification": {
    "legibility": "<one sentence>",
    "completeness": "<one sentence>",
    "correctness": "<one sentence>"
  }
}
\end{verbatim}

\end{prompt}

\subsection{Render}

\subsubsection{System prompts}

\begin{prompt}[The system prompts of LLM-as-a-judge on Render]
You are a strict and literal visual evaluator.

Your task is to compare TWO images that both contain text.

You must ONLY judge based on the VISIBLE text in the images.

Rules:

- Do NOT infer missing text.

- Do NOT correct errors.

- Do NOT use any external knowledge.

- Do NOT judge whether the text content itself is correct or meaningful.

- The provided caption is for reference only and MUST NOT be used to infer missing text.

- If text is missing, unclear, altered, or unreadable, you MUST penalize it.

- If you are uncertain, choose the LOWER score.

You are NOT a reasoning assistant.

You are a consistency checker.
\end{prompt}

\subsubsection{User prompts}
\begin{prompt}[The user prompts of LLM-as-a-judge on Render]
You are given:

- A target caption (for reference only):
<CAPTION>

- Image A: a model-generated image.

- Image B: a reference image.

Both images contain text.

The task is to render the target caption as white text on a black background.

Your job is to compare Image A with Image B and evaluate ONLY the visible text.

IMPORTANT:

- The target caption is provided ONLY for context and MUST NOT be used to infer, complete, or correct any missing or unreadable text.

- You MUST base your judgment strictly on the visible text in the images.

- Do NOT guess or infer any missing content.

Please evaluate the following aspects using discrete integer scores from 0 to 5:

1. Text Legibility (0–5)

Is the text in Image A clearly readable by a human?

Score:

- 5: Completely clear and sharp; all text is easily readable without effort.

- 4: Mostly clear; minor blur or artifacts but all text is readable.

- 3: Readable with noticeable effort; blur, low contrast, or layout issues slow reading.

- 2: Partially readable; some words or characters are difficult or impossible to recognize.

- 1: Barely readable; only a few characters or fragments can be identified.

- 0: Unreadable; text cannot be recognized at all.

2. Text Completeness (0–5)

Is all text in Image B fully present in Image A?

Score:

- 5: Fully complete; all text is present with no missing parts.

- 4: Almost complete; only negligible parts are missing.

- 3: Partially complete; main content is present but some sections are missing.

- 2: Largely incomplete; significant portions are missing.

- 1: Barely complete; only a few words or fragments remain.

- 0: Completely missing; almost no text from Image B is present.

3. Text Correctness (0–5)

Does Image A contain the same textual content as Image B?

Score:

- 5: Text is exactly the same or strictly semantically equivalent.

- 4: Nearly equivalent; only trivial differences (e.g., punctuation, spacing, capitalization).

- 3: Mostly equivalent; core content matches but with noticeable wording or symbol differences.

- 2: Partially equivalent; some content matches, but keywords or structure are incorrect or altered.

- 1: Minimally related; only small fragments overlap.

- 0: Not equivalent at all or completely unrelated.

Rule:

If Image A contains extra text not present in Image B, correctness MUST NOT exceed 3.

IMPORTANT:

- Base your judgment ONLY on visible text.

- Do NOT assume correctness based on plausibility.

- Extra text in Image A should reduce correctness.

- If unsure, choose the lower score.

- Scores must be integers between 0 and 5. Do NOT use decimal values.

Return your judgment in the following JSON format ONLY:
\begin{verbatim}
{
  "correctness": <0-5>,
  "completeness": <0-5>,
  "legibility": <0-5>,
  "justification": {
    "legibility": "<one sentence>",
    "completeness": "<one sentence>",
    "correctness": "<one sentence>"
  }
}
\end{verbatim}

\end{prompt}

\section{Lowless Visualization}
\begin{figure}
  \centering
   \includegraphics[width=0.7\linewidth]{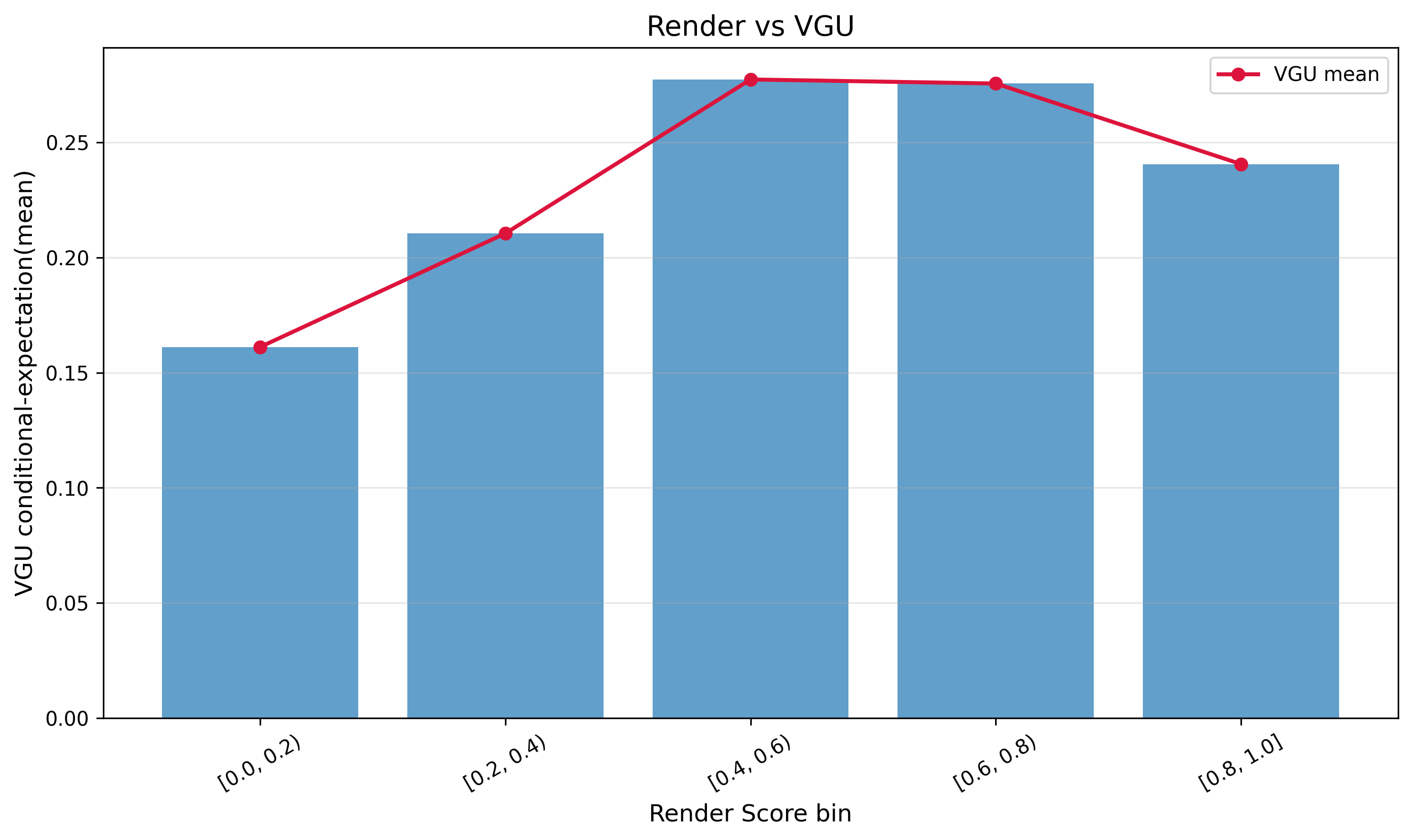}
   \caption{Trend of conditional expectation of VGU values across Render score bins. The red LOWESS curve illustrates how the average VGU value changes with Render scores, highlighting a non-linear relationship that global correlation metrics may not capture.}
   \label{fig:lowless}
\end{figure}

As shown in \Cref{fig:lowless}, in addition to the dependency analysis presented in \Cref{sec:method}, we provide further visualization of the conditional expectation between VGU and Render.

\end{document}